\documentclass[10pt,twocolumn,letterpaper]{article}

\usepackage{cvpr}              % To produce the CAMERA-READY version
\usepackage{graphicx}
\usepackage{amsmath}
\usepackage{amssymb}
\usepackage{booktabs}
\usepackage[accsupp]{axessibility}

\usepackage[pagebackref,breaklinks,colorlinks]{hyperref}

\usepackage[capitalize]{cleveref}
\crefname{section}{Sec.}{Secs.}
\Crefname{section}{Section}{Sections}
\Crefname{table}{Table}{Tables}
\crefname{table}{Tab.}{Tabs.}

\def\cvprPaperID{*****} % *** Enter the CVPR Paper ID here
\def\confName{CVPR}
\def\confYear{2022}

\begin{document}

%%%%%%%%% TITLE - PLEASE UPDATE
\title{Symmetric Network with Spatial Relationship Modeling for \\Natural Language-based Vehicle Retrieval}

\author{Chuyang Zhao\thanks{Equal Contribution} \quad Haobo Chen\footnotemark[1] \quad Wenyuan Zhang \quad Junru Chen \\ Sipeng Zhang \quad Yadong Li  \quad Boxun Li  \\
MEGVII Technology \\
{\tt\small \{zhaochuyang,chenhaobo,zhangwenyuan,chenjunru,zhangsipeng,liyadong,liboxun\}@megvii.com}}

\maketitle

%%%%%%%%% ABSTRACT
\begin{abstract}
Natural language (NL) based vehicle retrieval aims to search specific vehicle given text description. 
Different from the image-based vehicle retrieval, NL-based vehicle retrieval requires considering not only vehicle appearance, but also surrounding environment and temporal relations. 
In this paper, we propose a \textbf{S}ymmetric Network with \textbf{S}patial Relationship \textbf{M}odeling (SSM) method for NL-based vehicle retrieval. 
Specifically, we design a symmetric network to learn the unified cross-modal representations between text descriptions and vehicle images, where vehicle appearance details and vehicle trajectory global information are preserved.
Besides, to make better use of location information, we propose a spatial relationship modeling methods to take surrounding environment and mutual relationship between vehicles into consideration. 
The qualitative and quantitative experiments verify the effectiveness of the proposed method.
We achieve 43.92\% MRR accuracy on the test set of the 6th AI City Challenge on natural language-based vehicle retrieval track, yielding the 1st place among all valid submissions on the public leaderboard. The code is available at \url{https://github.com/hbchen121/AICITY2022_Track2_SSM}.
% Besides, SSM utilize spatial relationship modeling to take surrounding environment and mutual relationship between vehicles into consideration. The qualitative and quantitative experiments verify the effectiveness of the proposed method. Our proposed method has achieved 43.92\% MRR accuracy on the test set of the 6th AI City Challenge on natural language-based vehicle retrieval track, yielding the 4th place on the public leaderboard. The code will be available at \url{https://github.com/hbchen121/AICITY2022_Track2_SSM}.

\end{abstract}
\section{Introduction}
Natural Language (NL) based vehicle retrieval aims to find specific vehicles given text descriptions, which is a vital part of the intelligent transportation and urban surveillance systems.
Existing vehicle retrieval systems are generally based on image-to-image matching, \textit{i.e.,} the vehicle re-identification.
It requires providing a query image to retrieve the same vehicle in the image gallery.
However, in practical applications, the query image may not be available and we only know the rough description of the target vehicle.
This leads us to an urgent need for an NL-based vehicle retrieval system.

% Compared with image queries, natural language descriptions are easier to obtain and easier to modify, so it has higher flexibility. 
Compared with image queries, natural language text descriptions are easier to obtain and modify, resulting in higher flexibility.
However, because text belongs to a different modality from the image, it brings great challenges to retrieval.
Specifically, natural language-based vehicle retrieval is a cross-modal matching that aims to learn a transferable model between visual and language. 
DUN \cite{sun2021dun} proposes a dual-path network to align vehicle features in video and text embeddings supervised by the circle loss \cite{CircleLoss}.
But they ignore motion information in matching, which is conducive to distinguishing similar vehicles with different motion states.
Different from DUN, CLV \cite{bai2021connecting} utilize a symmetric InfoNCE loss \cite{xie2018rethinking} to learn cross-modal representation like CLIP \cite{radford2021learning}.
They generate a global motion image for each track, retaining not only vehicle appearance information.
In addition, the subject descriptions in the text are enhanced, improving the impact of vehicle appearance information.
However, their motion maps suffer from a loss of vehicle appearance information, which degrades model performance.
% In recent years, more and more works focus on cross-modal representation learning. 
% Inspired by previous works, we perform the symmetric InfoNCE loss like CLIP \cite{radford2021learning} and pair-wise Circle loss \cite{sun2020circle} to maximize the cosine similarity of image and text representations.

Inspired by these methods, we propose a \textbf{S}ymmetric Network with \textbf{S}patial Relationship \textbf{M}odeling (SSM) approach to learn the visual and linguistic representations for NL-based vehicle retrieval.
The symmetric network can learn both the vehicle internal characteristic (\textit{e.g.} vehicle type, color, and shape) and the vehicle external characteristic (\textit{e.g.} motion state and surrounding environment) simultaneously.
% a dual stream network to learn both the vehicle internal characteristic (e.g. vehicle type, color and shape) and the vehicle external characteristic (e.g. motion state and surrounding environment). 
% More concretely, we adopt two visual backbone to extract the vehicle internal feature and the vehicle external feature respectively.
More concretely, we adopt one visual encoder and text encoder to extract the vehicle appearance feature and the vehicle text embedding, respectively, which is optimized by the symmetric InfoNCE loss and pair-wise Circle loss.
Symmetrically, another visual encoder and text encoder are employed to learn the representations of vehicle external characteristics, respectively.
The visual encoder can be the EfficientNet B2 \cite{xie2018rethinking} and the text encoder can be RoBERTa \cite{liu2019roberta}.
Both internal and external features are fused to learn robust features of the vehicle track.
% We also use the BERT network \cite{} to extract the language embedding of the sentences describing the internal and external features of the vehicle, respectively and then use two fully connected networks to project their embeddings into the same space as the visual embeddings. 
We also propose a hard text sample mining method to distinguish similar texts, where texts from other views are utilized.
In addition, by analyzing the ranking visualization, we found that the model can hardly learn the visual feature of the surrounding environment and relationship of multiple vehicles. 
To tackle this problem, we design a spatial relationship modeling approach to enhance tracklet information, improving the performance of our method.
% we use tracklet information to augment visual feature which greatly improve the performance of our retrieval system. 
We has achieved 43.92\% MRR accuracy on the test set of
the 6th AI City Challenge on natural language-based vehicle retrieval track, yielding the 1st place among all valid submissions on the public leaderboard.

\section{Related Work}

\subsection{Natural Language Based Video Retrieval}
Natural language based video retrieval aims to find the corresponding video given natural language description. There are increasing interest in this area. \cite{li2021x,qi2021semantics,han2021fine,gabeur2022masking}
Most of the existing methods are based on representation learning, which try to make the feature representation of the corresponding text and video similar. These methods usually use a language model such as LSTM\cite{hochreiter1997long} or BERT, etc. \cite{devlin2018bert} to extract the language feature and use a visual feature backbone such as ResNet \cite{he2016deep} or C3D \cite{carreira2017quo}, etc. to extract the visual feature.
Zhang, et al. \cite{zhang2018cross} use hierarchical sequence embedding (HSE) to embedding sequential data of different modalities into hierarchically semantic spaces with correspondence information. Antoine, et al.\cite{miech2018learning} propose a Mixture-of-Embedding-Experts (MEE) model with ability
to handle missing input modalities during training. Dong, et al.\cite{dong2019dual} proposing a dual
deep encoding network that encodes videos and queries into
powerful dense representations of their own. Our method not only utilizes representation learning to learn transferable visual and text feature, but also utilize spatial relationship modeling to capture the surrounding environment and mutual relationship in vehicle retrieval task.

\subsection{Vehicle Re-identification}
In the last decade, vehicle ReID has achieved considerable progress, especially for deep learning based methods \cite{FACT,he2019part,zhuge2020attribute,chen2021vehicle}.
Different from the person ReID methods \cite{sun2018beyond,zheng2022template}, it is still a challenging task due to the similar appearance and viewpoint variations problems.
To deal the two problems, some approaches \cite{OrientationVeri,he2019part,Meng2020} focus on subtle details to learn discriminative local features.
For instance, OIFE \cite{OrientationVeri} proposes 20 keypoints and raises an orientation invariant feature embedding module to emphasize regions with discriminative information.
Part-regularized \cite{he2019part} trains a detector to focus on local regions and acquires local features to distinguish similar vehicles.
VANet \cite{Chu2019} adopts a view predictor and a modified triplet loss to generate viewpoint-aware features.
In addition, many loss functions are applied to address the above challenges, including representation learning loss functions and metric learning loss functions.
representation learning losses \cite{LSoftmax,NormSoftmax,AMSoftmax,CosFace,ArcFace} acquire feature representations by classification.
Unlike them, metric learning losses \cite{LeCun2005,triplet2017,CircleLoss} directly optimize the similarity score of image pairs and are term as pair-wise loss, where margins are generally added between positive pairs and negative pairs to increase the distance.
In this paper, we learn the unified cross-modal representation through the pair-wise losses.

\section{Method}

\subsection{Overview}

The pipeline of our method is illustrated in Fig. \ref{fig:framework}.
The symmetric network consists of four branches, where the upper two branches are mainly used to acquire local visual and language representations of vehicle appearance, and the lower two branches are employed to learn the global information of the vehicle, including the motion state and the environment in which it is located.
The two kinds of representations are fused together to generate the comprehensive feature representations of the vehicle track.
We apply the Symmetric InfoNCE Loss \cite{oord2018representation} and Circle Loss \cite{sun2020circle} to connect the representations of the two modalities of the text and the image, ensuring that they are projected into a unified representation space.
% 为了进一步强化非车辆外观信息的track信息的作用（motion，intersection等），减小相似车辆不同track对模型的影响，防止车辆外观信息统治模型，我们
To further emphasize the motion and environment information, we construct triples to distinguish text descriptions from different views of the same vehicle.
In addition, a spatial relationship modeling module is designed to use location information to constrain retrieval results in the retrieval process.

\subsection{Data Augmentation}

\subsubsection{Image Augmentation}

\begin{figure}[t]
\begin{center}
% \vspace{-5mm}
\includegraphics[width=1.\columnwidth]{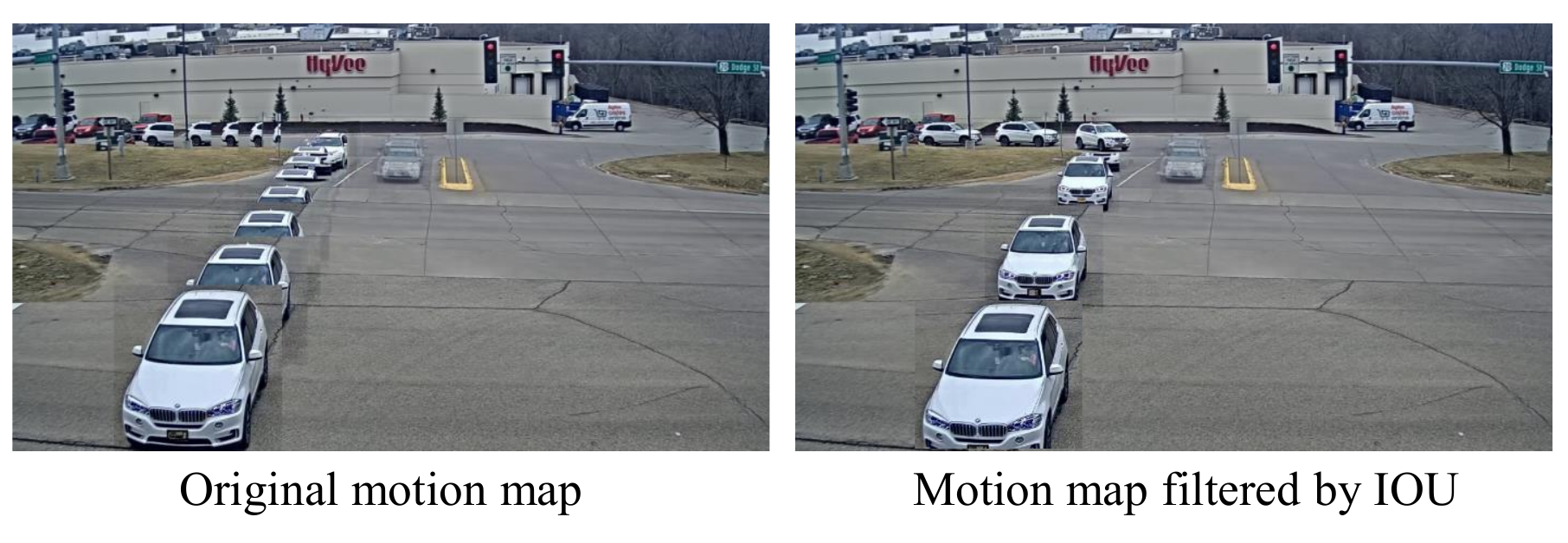}
\end{center}
% \vspace{-7mm}
\caption{
Comparison of motion maps.
We use cropped car image files to generate IOU-filtered motion maps that include more vehicle information.
}
\label{fig:iou}
\end{figure}
% \vspace{-3mm}

Similar to \cite{bai2021connecting}, we augment the motion image by pasting cropped vehicle images from the tracking frames to the background image. 
% We compute the mean value of all frames from the same camera to generate the background image, which can be formulated as:
The background image is generated by computing the mean value of all frames from the same camera, which can be formulated as:
$$
B_c = \frac{1}{N_c}\sum_i^{N_c} {F_i}
$$
where $B_c$ is the background image of the $c_{th}$ camera, $N_c$ is the number of frames taken by the $c_{th}$ camera, and $F_i$ is the $i_{th}$ frame. 

The motion image is generated by pasting cropped vehicle images from the same track on the corresponding background image.
As shown in the left of Fig. \ref{fig:iou}, too close consecutive frames may occlude adjacent frames on the motion image.
To avoid this problem, we compute the Intersection Over Union (IOU) of adjacent frames and ignore frames whose IOU is larger than the threshold with already pasted frames.
The threshold is set to 0.05 in our method.
% To keep the consecutive frames not too close, which may occlude adjacent frames on the motion image, we compute the IOU of adjacent frames and ignore frames whose IOU is larger than the threshold with already pasted frames. We set the IOU threshold to 0.05 in our experiments.

% 补一个公式化的句子
% 补一组设置 IOU threshold 和不设置的对比图

\subsubsection{Text Augmentation}
% 介绍存在的问题
% 提出了增强的方式

\begin{figure}[t]
\begin{center}
% \vspace{-5mm}
\includegraphics[width=1.\columnwidth]{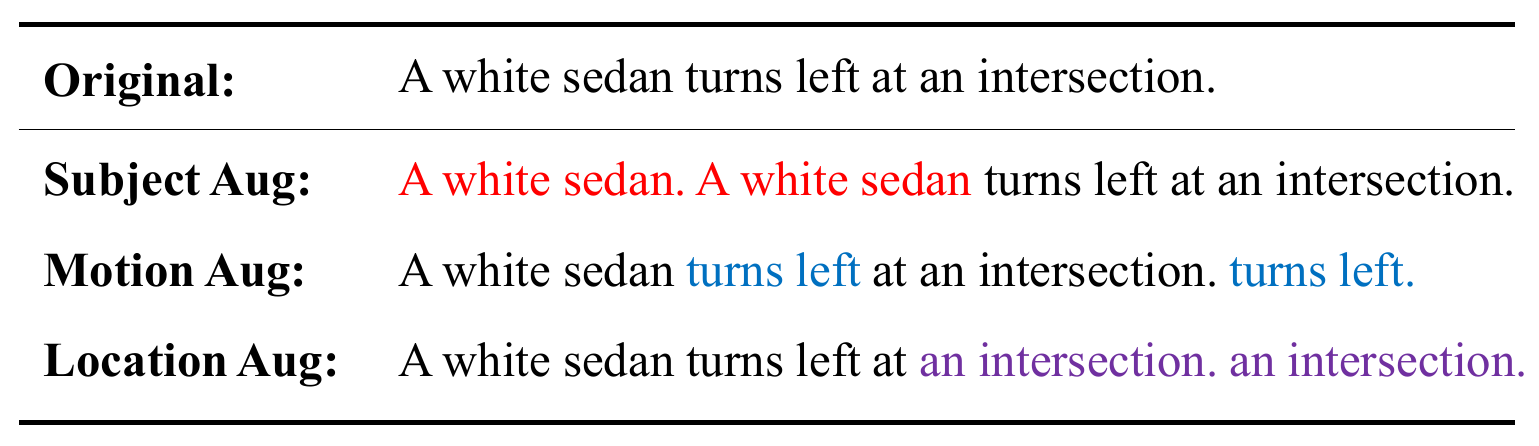}
\end{center}
% \vspace{-7mm}
\caption{
Examples of augmentation in text descriptions.
}
\label{fig:text}
\end{figure}
% \vspace{-3mm}

Most text descriptions contain vehicle appearance information, motion information and location information. 
To make the model learn a specific information better,
we adopt subject augmentation, motion augmentation and position augmentation respectively to emphasize different textual information, in which the respective text descriptions are repeated as shown in Fig. \ref{fig:text}.
\\
\hspace*{\fill} 
\\
\textbf{Subject augmentation.} To enhance the appearance information, we need to repeat description about the vehicle appearance. Because the vehicle appearance description usually appears as the first noun phase of the sentence, we "spacy "\footnote{\tiny\url{https://spacy.io/}} to extract all noun phrases and put the first one to the beginning of the sentence.
\\
\hspace*{\fill} 
\\
\textbf{Motion augmentation.} To enhance the motion information, we extract keywords which describe motion information and then repeat them. 
For motion information, We assume that there are only three motion types, \textit{i.e.} turn left, turn right and go straight. 
If keywords ``turn left" or ``turn right" exist in any text description of one track's text description, we denote the motion type of this track is turning left or right. 
If none of these two keywords exist in the text description, we denote the motion type of this track is going straight. 
We append the motion type keyword ``left", ``right" or ``straight" to the beginning of all text description of the track to emphasize the motion information. 
\\
\hspace*{\fill} 
\\
\textbf{Location augmentation.} For location enhancement, We only distinguish whether a location described by a track is an intersection. 
So we look into the text descriptions of the track and check whether ``intersection" exists in any of the text description to determine whether the described location is an intersection. 
We append ``intersection" to the beginning of all text description of the track if the track is located in the intersection.

\begin{figure*}[t]
\begin{center}
% \vspace{-5mm}
\includegraphics[width=2.1\columnwidth]{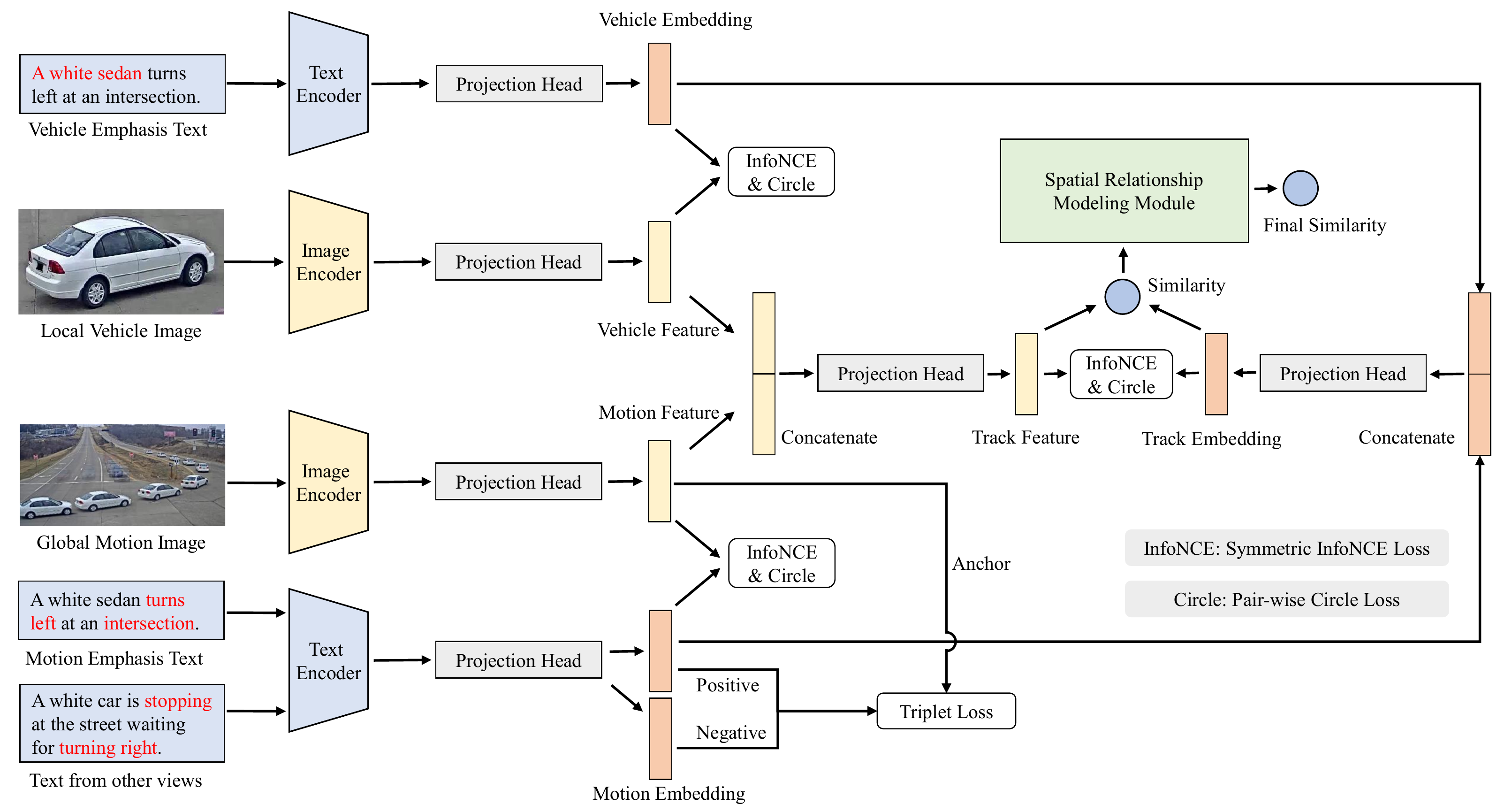}
\end{center}
% \vspace{-7mm}
\caption{
The main pipeline of the proposed method. 
There are four branches in our symmetric network.
The upper two branches focus on the internal characteristic of vehicles, where the subject in text is enhanced and the local vehicle image is processed.
Symmetrical with it are the lower two branch, in which the external characteristic are acquired.
Both kinds of representations are used together to retrieve vehicle images with textual descriptions.
}
\label{fig:framework}
\end{figure*}
% \vspace{-3mm}

\subsection{Visual and Linguistic Representation Learning}

\subsubsection{Symmetric Network}

The task of natural language based vehicle retrieval refers to measuring the semantic similarity between a vehicle video and a language description.
Generally speaking, humans pay more attention to unique information when describing vehicle videos, such as vehicle appearance, shape and other characteristics (termed as internal characteristics), as well as external characteristics such as motion state and surrounding environment (termed as external characteristics).
We construct a dual-stream network to focus the internal and external characteristics, respectively.
\\
\hspace*{\fill} 
\\
\textbf{Vehicle internal characteristic learning.}
% Specifically, 
As the upper two branches shown in Fig. \ref{fig:framework},
to enhance the internal characteristic in vehicle images, we crop the local vehicle image from the frame of the video and feed them into the image encoders.
The encoders are EfficientNet B2 \cite{xie2018rethinking} or ibn-ResNet101-a \cite{pan2018two} pretrained on ImageNet \cite{deng2009imagenet}.
% The vehicle image is cropped from the frame of the video, and is fed into the image encoders, which is EfficientNet B2 \cite{xie2018rethinking} or ibn-ResNet101-a \cite{pan2018two} pretrained on ImageNet \cite{deng2009imagenet}.
In addition, we employ a pretrained RoBERTa \cite{liu2019roberta} as text encoder to learn the text embedding of vehicles.
Its input text is augmented by subject emphasis to acquire vehicle embedding with more appearance and shape information.
Following \cite{bai2021connecting}, we adopt projection heads to map visual feature and text embedding into a unified representation space, which is formulated as:
\begin{equation}
f_{i}=g_{i}\left(h_{i}\right)=W_{2} \sigma\left(B N\left(W_{1} h_{i}\right)\right),
\end{equation}
\begin{equation}
f_{t}=g_{t}\left(h_{t}\right)=W_{2} \sigma\left(L N\left(W_{1} h_{t}\right)\right),
\end{equation}
where $h_i$ is the visual feature extracted by the image encoder and $h_t$ is the linguistic embedding extracted by the image encoder.
BN denotes the Batch Normalization (BN) layer. 
LN is a Layer Normalization (LN) layer.
$\sigma$ is a ReLU activation layer.
\\
\hspace*{\fill} 
\\
\textbf{Vehicle external characteristic learning.}
Similar to the network structure in internal features learning, we adopt a image encoder and a text encoder in the lower two branches to learning global motion feature and embedding, respectively.
The input of global image encoder is the global motion image generated by multiple frames of images in the video, which is beneficial to learn more action and environmental information.
The motion emphasis or location emphasis augmented text description is processed by the global text encoder to extract more vehicle external information.
The obtained global motion feature and motion embedding are projected into the same representation space as well.
\\
\hspace*{\fill} 
\\
\textbf{Vehicle comprehensive characteristic learning.}
We concatenate the local and global image features (text embedding) to fuse information at different granularities.
The fused visual and linguistic representations are projected into the same space by the projection heads.
Then in the inference process, we only utilize the fused track features.

\subsubsection{Loss functions}

In the unified feature representation space containing two modal information, we perform the symmetric InfoNCE loss \cite{xie2018rethinking} like CLIP \cite{radford2021learning} and pair-wise Circle loss \cite{sun2020circle} to maximize the cosine similarity of image and text representations.

Given $N$ images and $N$ text descriptions, we can acquire $N$ visual features $f^{img}_i$ and text embedding $f^{text}_i$, where $f^{img}_i$ and $f^{text}_i$ have the same label and become positive pairs with each other, \textit{i.e.}, $\langle f^{img}_i, f^{text}_i\rangle$. 
These features has different labels are negative pairs of each other, including same-modal negative pairs $\langle f^{img}_i, f^{img}_{j \ne i}\rangle$ and cross-modal negative pairs $\langle f^{img}_i, f^{text}_{j \ne i}\rangle$.

The symmetric InfoNCE loss maximize the cosine similarity of the positive pair and minimize the similarity of cross-modal negative pairs.
It contains image-to-text optimization and text-to-image optimization for one positive pair.
The image-to-text loss is as follows:
\begin{equation}
\mathcal{L}_{i2t}=\frac{1}{N} \sum_{i=1}^{N}-\log \frac{\exp(\cos ( f^{img}_i,f^{text}_i) / \tau)}{\sum_{j=1}^{N} \exp(\cos ( f^{img}_i,f^{text}_j) / \tau)},
\end{equation}
In addition, the text-to-image loss is:
\begin{equation}
\mathcal{L}_{t2i}=\frac{1}{N} \sum_{i=1}^{N}-\log \frac{\exp(\cos ( f^{text}_i,f^{img}_i) / \tau)}{\sum_{j=1}^{N} \exp(\cos ( f^{text}_i,f^{img}_j) / \tau)},
\end{equation}
where $\tau$ is the temperature learnable parameter and the cosine similarity is calculated by:
\begin{equation}
\cos (f_i, f_j) = \frac{f_i \cdot f_j}{||f_i|| \cdot ||f_j||}.
\end{equation}
Then, the symmetric InfoNCE loss is as follows:
\begin{equation}
\mathcal{L}_{INCE} = \mathcal{L}_{i2t} + \mathcal{L}_{t2i}.
\end{equation}

Different from the symmetric InfoNCE loss, Circle loss minimize the similarity of all negative pairs.
Denote the positive pair and negative pairs as $s_p$ and $s_n$, respectively.
Circle loss is defined as follows:
\begin{equation}
\tiny
\mathcal{L}_{{circle}}=\log \left[1+\sum_{j=1}^{L} \exp \left(\gamma \alpha_{n}^{j}\left(s_{n}^{j}-\Delta_{n}\right)\right) \sum_{i=1}^{K} \exp \left(-\gamma \alpha_{p}^{i}\left(s_{p}^{i}-\Delta_{p}\right)\right)\right]
\end{equation}
where the $K=1$ and $L=2(N-1)$, $\Delta_{p}$ and $\Delta_{n}$ are the intra-class and inter-class margins, respectively.
$\alpha_{p}$ and $\alpha_{n}$ are calculated as:
\begin{equation}
\small
\begin{cases}
\alpha_{p}^{i}=\left[O_{p}-s_{p}^{i}\right]_{+} \\
\alpha_{n}^{j}=\left[s_{n}^{j}-O_{n}\right]_{+}
\end{cases},
\end{equation}
in which $O_p$ and $O_n$ are optimums of $s_p$ and $s_n$, respectively.
These parameters are set $O_{p} = 1 + m$, $\Delta _{p} = 1 - m$, $O_{n} = -m$ and $\Delta _{n} = m$, respectively.

The symmetric InfoNCE loss and Circle loss are enforced on all of three kinds of representations, which preserves information of different levels can be learned.

\subsubsection{Hard Text Samples Mining}

% nlp other views

Although the non-appearance information is enhanced in external feature learning,  our model still learns a lot of vehicle internal information and makes mistakes when discriminating difficult samples.
For instance, when the same car appears in two different videos, the text descriptions of the two videos will have similar subjects as show in Fig. \ref{fig:framework}, which will degrade the retrieval performance.

To address this problem, we implement hard sample mining by composing triples with different view  descriptions of the same vehicle and the current view motion image.
More concretely, given $N$ global motion images, $N$ text descriptions and $N$ text descriptions from other views, we have $N$ triplets $\langle f^{img}_i, f^{text}_i, f^{text\_v}_i\rangle$, where the anchor is $f^{img}_i$, the positive feature is $f^{text}_i$ and the negative feature is $f^{text\_v}_i$.
Then we minimize the Triplet loss \cite{hoffer2015triplet} to push away the $f^{text\_v}_i$ from $f^{img}_i$:
\begin{equation}
\small
\mathcal{L}_{triple}=\frac{1}{N} \sum_{i=1}^{N}\left[ \cos ( f^{img}_i,f^{text\_v}_i) - \cos ( f^{img}_i,f^{text}_i) + m\right]_{+} \\
\end{equation}
in which $m$ is the margin.

\subsection{Spatial Relationship Modeling}
\subsubsection{Long-distance Relationship Modeling}
% 1. 总结
% 2. Insight
% 3. 做法
Long-distance relationship modeling means that we build positional relationships between text and images, \textit{e.g.}, intersection prediction.
Then the text and images whose positions are both intersections are given greater similarity
Specifically, we distinguish whether a location is intersection.
To increase the similarity of the text and image feature if they all describe the intersection, We calculate the location similarity between the text and the image and then add the similarity score to the final similarity matrix.

We found that the pictures in the dataset are all taken by cameras in fixed locations, so we can determine whether the location of an image is at the intersection by checking the location of its corresponding camera. We found if a camera is located in the intersection, some vehicles captured by this camera will stop for a while to wait the traffic light. 
We can calculate the location of the vehicle in each frame using the bounding box:
\begin{equation}
(x_i, y_i) = (left_i+\frac{1}{2}width_i, top_i + \frac{1}{2}height_i)
\end{equation}
where $left_i$  and $top_i$ are the left-top corner coordinate of the $i_{th}$ frame's bounding box, $width_i$ and $height_i$ are the width and height of the $i_{th}$ frame's bounding box.

We can infer the movement of the vehicle through the coordinates in each frames. If movement distance of the vehicle in consecutive $n$ frames are all zero, we consider the car to be at the intersection and we label the corresponding camera to be located at the intersection. We can get the visual location vector by:
\begin{equation}
  loc_v = 
    \begin{cases}
        (0, 1)^T,& \quad c_v \in \{c | c ~\text{is at intersection}\} \\
        (1, 0)^T,& \quad \text{otherwise}
    \end{cases}       
,\end{equation}
where $c_v$ is the camera of track $v$, $\{c | c ~\text{is at intersection}\}$ is the collection of cameras which are located in the intersection.

We can determine whether a sentence is describing an intersection by checking whether keyword ``intersection" exists in the sentence. We can get the text location vector by:
\begin{equation}
  loc_s = 
    \begin{cases}
        (0, 1)^T,& \quad \text{``intersection"} \subseteq  s \\
        (1, 0)^T,& \quad \text{otherwise}
    \end{cases}       
,\end{equation}

After we get the visual location embedding and text location embedding, we can calculate the similarity of them by the dot product. Assume we have $n$ text queries and $m$ visual tracks, we can get a matrix $S_l \in \mathbb{R} ^ {n\times m}$ representing the location similarity between each query and track. 

\subsubsection{Short-distance Relationship Modeling}
In the queries, there are lots of sentences describe the relationship of more than one vehicles. 
However, Our model can not learn the relationship between multiple vehicles, which are very close in the video frame.
To utilize this information, we perform relationship augmentation to make the proper text-visual pair more similar, termed short-distance relation modeling.

More concretely, we use `spacy' to extract all noun-phases in the sentence and keep all phases describing vehicles. Then we use the vehicle description branch to extract the language embeddings of all the vehicle descriptions. We then use the provided detection file to extract the bounding boxes of all cars in the frame. We randomly select several frames in a track and then extract all detected cars in these frames. We use the cropped visual branch to extract the visual embedding for all cars in these frames. If a text query $q$ describes relationship between vehicles $v_1$ and $v_2$, we can calculate the similarity between $v_2$ and all detected vehicles in track $t$ and take the maximum value as the relationship similarity between $q$ and $t$. Assume we have $n$ text queries and $m$ visual tracks, we can get a matrix $S_r \in \mathbb{R} ^ {n\times m}$ representing the relationship similarity between each query and track.

The final similarity matrix $S_{final}$ is formulated as the sum of the feature similarity matrix $S$, location similarity matrix $S_l$ and relationship similarity matrix $S_r$:

\begin{equation}
S_{final} = S + \alpha S_l + \beta S_r
\end{equation}
$\alpha$ and $\beta$ are hyper-parameters. We set $\alpha=1$ and $\beta=0.2$ in our experiments.
\section{Experiments}

\subsection{Datasets and Evaluation Settings.}

\noindent\textbf{Datasets.} 
CityFlow-NL \cite{Feng21CityFlowNL} dataset consist of 666 vehicles contains 3028 vehicle tracks collected from 40 cameras, of which 2155 vehicle tracks were used for training.
Each track was annotated with three natural language descriptions.
In addition, multiple natural language descriptions from other views of the vehicle in this track are collected in this track.
The remaining data were adopted to evaluate the proposed method.
\\
\hspace*{\fill} 
\\
\textbf{Evaluation.} The vehicle retrieval by NL descriptions task 
generally utilize the Mean Reciprocal Rank (MRR) as the standard metrics, which is also used in \cite{voorhees1999trec}.
It is formulated as:
\begin{equation}
MRR=\frac{1}{|Q|} \sum _{i=1}^{|Q|}\frac {1}{rank_{i}},
\end{equation}
where $|Q|$ is the number of set of text descriptions. $rank_{i}$ refers to the rank position of the right track for the $i_{th}$ text description. 

\subsection{Implement details}

Both input local and global images are resized to 228 $\times$ 228.
We do not adopt any image augmentation methods to augment the data, including random horizontal flip and random clips, which can keep sports information such as motion unchanged.
We set the batch size to 64 for each input, \textit{i.e.,} vehicle local images, vehicle emphasis texts, global motion images, motion emphasis texts and text from other views.
The total batch size is 320.
Two text encoder is frozen in the training process. 
Our model is trained for 400 epochs employing the AdamW \cite{loshchilov2017adamw} optimizer with weight decay (1e-2).
The learning rate starts from 0.01 and works as a warm-up strategy for 40 epochs. 
A step delay scheduler is adopted to decay the learning rate every 80 epochs. 
The $\gamma$ and $m$ in circle loss are set to 48 and 0.35, respectively.
The $m$ in triplet loss is set to 0.
We train the model with different configures and ensemble them in final similarity calculating.

\subsection{Performance on CityFlow-NL}

\begin{table}
% \small
\centering
\caption{Ablation Study Results of our method.}
\begin{tabular}{l c}
\hline
Method & MRR\\
\hline
Baseline & 0.3252 \\
Baseline + Motion Aug & 0.3362 \\
Baseline + Location Aug & 0.3305 \\
Baseline + Other views & 0.3405 \\
Baseline + Circle loss & 0.3548 \\
Ensemble &  0.4207 \\
Ensemble + Spatial Modeling & 0.4392 \\
\hline
\end{tabular}
\label{tab:ablation}
\end{table}

\begin{table}
% \small
\centering
\caption{Competition results of AI City Natural Language-Based Vehicle Retrieval Challenge.}
\begin{tabular}{c|c|c}
\hline
Rank & Team Name & MRR\\
\hline
% 1 & Must Win & 0.6606 \\
% 2 & Thursday & 0.5251 \\
% 3 & HCMIU-CVIP & 0.4773 \\
1 & \textbf{MegVideo} & \textbf{0.4392}  \\
2 & HCMUS & 0.3611 \\
3 & P \& L & 0.338 \\
4 & Terminus-AI & 0.3320 \\
5 & MARS\_WHU & 	0.3205 \\
6 & BUPT\_MCPRL\_T2 & 0.3012 \\
7 & folklore & 0.2832 \\
\hline
\end{tabular}
\label{tab:rank}
\end{table}

We conduct ablation studies with different configures of our method.
The results are illustrated in Table \ref{tab:ablation}.
The ``Baseline'' denotes the symmetric network with subject augmentation and symmetric InfoNCE loss.
The ``other view`` refers to the hard samples mining approach.
It can be observed that each part learns different characteristics and brings a slight performance boost, respectively.
In addition, the model ensemble integrates different information and achieves a great improvement.

% We present the valid results of top-7 team in the public leaderboard, 
We present the top 7 valid results in the public leaderboard,
as shown in Table \ref{tab:rank}.
Our team (MegVideo) achieve a MRR score of 0.4392, taking 1st highest among all valid submissions on the AI City Challenge 2022 Track 2.
\section{Conclusion}
In this paper, we propose a {S}ymmetric Network with {S}patial Relationship {M}odeling method for NL-based vehicle retrieval.
Firstly, we design a symmetric network to learn visual and linguistic representations.
Then a spatial relationship modeling methods is proposed to make better use of location information, . 
The qualitative experiments confirm the effectiveness of our method. It achieves 43.92\% MRR accuracy on the test set of the 6th AI City Challenge on natural language-based vehicle retrieval track, yielding the 1st place among all valid submissions on the public leaderboard.
\\
\hspace*{\fill} 
\\
\textbf{Broader Impact.} 
Our research can promote the practical application of vehicle retrieval technology, but the data used in the research may cause privacy violations. Therefore, we only use the data of the public dataset and mask the license plate to avoid increasing privacy leaks and other security issues.

%%%%%%%%% REFERENCES
{\small
\bibliographystyle{ieee_fullname}
\bibliography{egbib}
}

\end{document}